\pdfoutput=1
\documentclass[11pt]{article}

\usepackage[final]{acl}

\usepackage{times}
\usepackage{latexsym}

\usepackage[T1]{fontenc}
\usepackage[utf8]{inputenc}

\usepackage{microtype}

\usepackage{inconsolata}

\usepackage{graphicx}
\usepackage{amsmath}
\usepackage{multirow}
\usepackage{booktabs}  % 让表格更美观
\usepackage{pifont}    % 提供 \xmark (×)
\newcommand{\cmark}{\ding{51}}%
\newcommand{\xmark}{\ding{55}}%
\usepackage{array}     % 控制表格列格式
\usepackage{boldline}  % 额外的加粗线条
\usepackage{tabularx}

\title{StructFlowBench: A Structured Flow Benchmark for Multi-turn Instruction Following}

\author{
        Jinnan Li$^{1,5}$ \quad Jinzhe Li$^{2}$ \quad Yue Wang$^{3}$ \quad  Yi Chang$^{1,4,5}$\footnotemark[1] \quad Yuan Wu$^{1}$\footnotemark[1] \\
        $^{1}$School of Artificial Intelligence, Jilin University \\
        $^{2}$College of Computer Science and Technology, Jilin University \\ 
        $^{3}$School of Information and Library Science, University of North Carolina at Chapel Hill \\
        $^{4}$Engineering Research Center of Knowledge-Driven Human-Machine Intelligence, MOE, China \\
        $^{5}$International Center of Future Science, Jilin University\\
        \{jnli23, lijz2121\}@mails.jlu.edu.cn, wangyue@email.unc.edu, \\
        yichang@jlu.edu.cn, yuanwu@jlu.edu.cn \\   
}

\begin{document}
\maketitle
\renewcommand{\thefootnote}{\fnsymbol{footnote}}
\footnotetext[1]{Corresponding authors}

\begin{abstract}
Multi-turn instruction following capability constitutes a core competency of large language models (LLMs) in real-world applications.
Existing evaluation benchmarks predominantly focus on fine-grained constraint satisfaction and domain-specific capability assessment, yet overlook the crucial structural dependencies between dialogue turns that distinguish multi-turn from single-turn interactions.
These structural dependencies not only reflect user intent but also establish an essential second dimension for the instruction following evaluation beyond constraint satisfaction.
To address this gap, we propose StructFlowBench, a multi-turn instruction following benchmark with structural flow modeling.
The benchmark defines an innovative structural flow framework with six fundamental inter-turn relationships.
These relationships introduce novel structural constraints for model evaluation and also serve as generation parameters for creating customized dialogue flows tailored to specific scenarios.
Adopting established LLM-based automatic evaluation methodologies, we conduct systematic evaluations of 13 leading open-source and closed-source LLMs. 
Experimental results reveal significant deficiencies in current models' comprehension of multi-turn dialogue structures.
The code is available at \url{https://github.com/MLGroupJLU/StructFlowBench}.
\end{abstract}

\section{Introduction}

The rapid advancement of large language models (LLMs) in multi-turn dialogue systems has elevated instruction-following capabilities to a pivotal research frontier in human-AI interaction~\cite{chang2024survey}.
Current evaluation methodologies bifurcate into two streams: multi-turn dialogue evaluations focusing on general capabilities~\cite{zheng2023judging,bai-etal-2024-mt,kwan-etal-2024-mt} and instruction-following analyses emphasizing fine-grained constraint compliance~\cite{jiang-etal-2024-followbench,he2024can,zhang2024cfbench}. 
More recent research has started to model the composition of intra-turn constraints~\cite{wen2024benchmarking}.

\begin{figure}[t!]
    \captionsetup{skip=0pt}
	\centering
	\includegraphics[width=\linewidth]{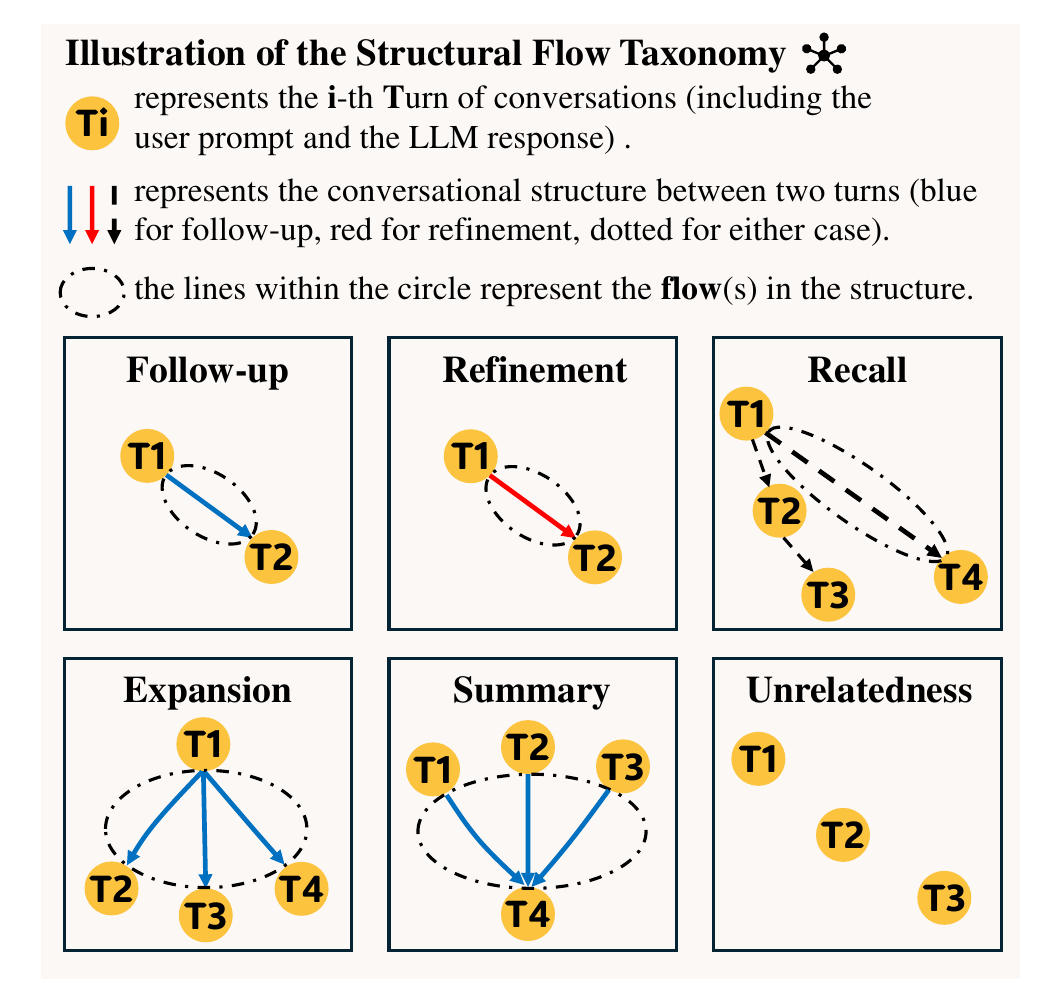}
	\caption{\textbf{The Structural Flow Taxonomy} includes six fundamental structures, each used to describe the inter-turn relationships in multi-turn dialogues. It can be applied to analyze any dialogue and generate specific structural flows.}
	\label{fig:intro}
\end{figure}
\vspace{-0.2em}

\begin{figure*}[t!]
    \captionsetup{skip=0pt}
	\centering
	\includegraphics[width=\textwidth]{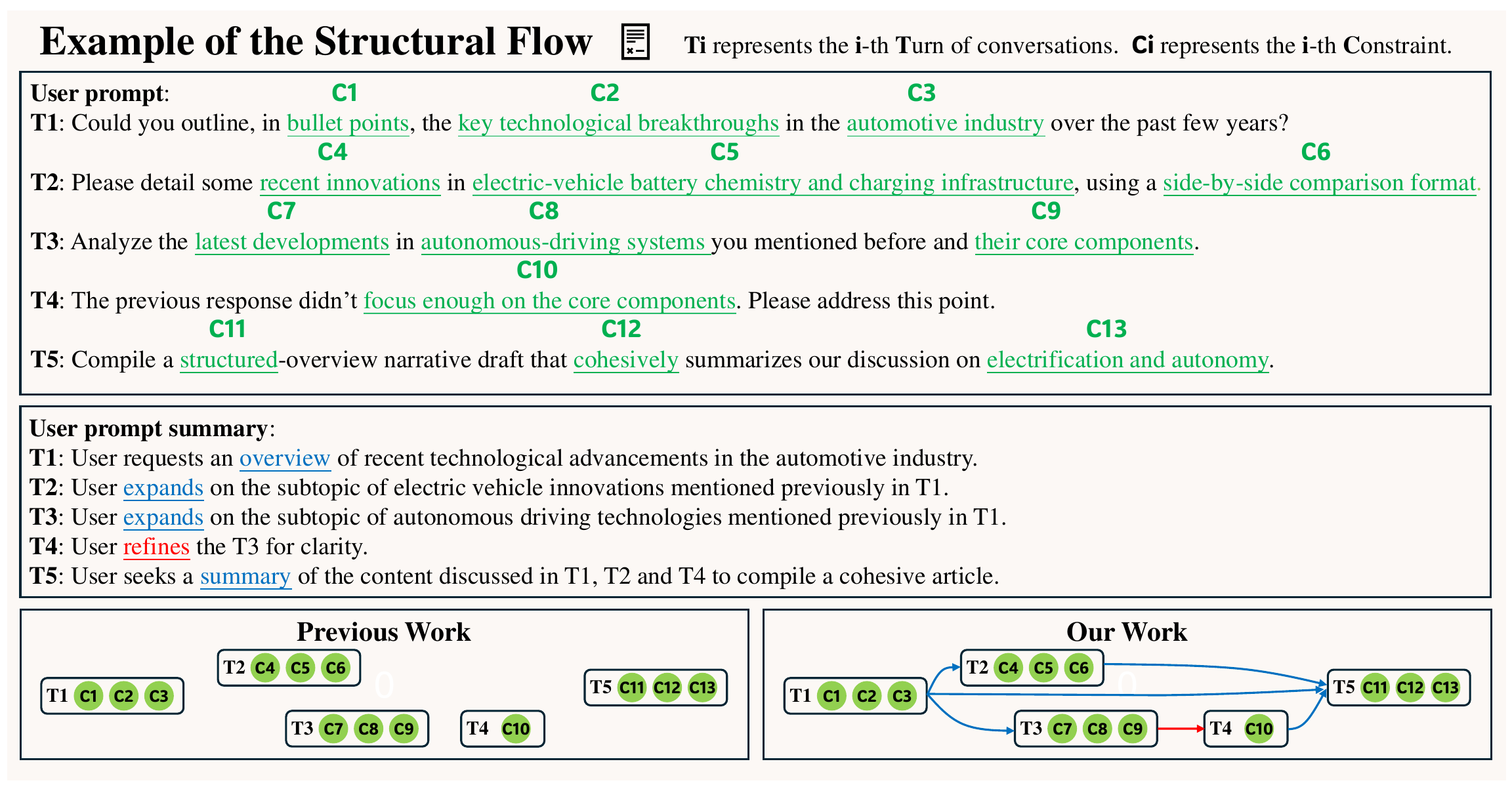}
	\caption{\textbf{An example of the Structural Flow} demonstrates how inter-turn relationships are represented using the proposed taxonomy, offering insight into the structure of real dialogues.}
	\label{fig:example}
\end{figure*}

However, current evaluation methodologies treat multi-turn dialogues merely as simple concatenations of single-turn interactions, neglecting users' planning and intentionality in extended conversations.
This leads to three critical limitations: (1) \textbf{Failure to model complex scenarios}: Multi-turn dialogue data constructed with simplistic linear thinking cannot accurately capture key characteristics of real-world complex conversations, such as logical coherence, user goal clarity, and natural transitions.
(2) \textbf{Methodological bias}: Single-turn evaluation strategies fragment inter-turn structural connections, overlooking multi-turn structural constraints.
(3) \textbf{Analytical deficiency}: Existing approaches overemphasize intra-turn constraint compliance while lacking a systematic framework to characterize dialogue structural flow.

To bridge these gaps, we introduce \textbf{StructFlowBench}, a novel instruction-following benchmark integrating a multi-turn structural flow framework. 
It consists of two key components: 1) \textbf{Dual-constraint evaluation system}, comprising two layers---intra-turn constraint evaluation and inter-turn constraint evaluation---offers a more comprehensive assessment of LLMs' multi-turn dialogue instruction-following capabilities. 
Building on existing intra-turn constraint assessments, we integrate five newly proposed structural constraints with eight intra-turn instruction constraints to capture the relationships between dialogue turns.
These structural constraints account for inter-turn dependencies, ensuring that models are evaluated not only on their ability to satisfy individual constraints but also on their capacity to maintain logical coherence across multiple turns.
2) \textbf{Six-category structural flow taxonomy}, encompassing six fundamental inter-turn relationships: \textit{Follow-up, Refinement, Recall, Summary, Expansion, Unrelatedness}.
Figure~\ref{fig:intro} illustrates the structural flow taxonomy, which is defined in detail in Section~\ref{sec:structural_flow_taxonomy}.
This taxonomy serves a tripartite function: (a) \textbf{Structural Diagnosis:} It enables a structured analysis of cross-turn structural rationality, helping to identify inconsistencies in dialogue flow and ensuring that model responses align with the expected discourse structure.
Figure~\ref{fig:example} provides an illustrative example, highlighting how structural flow modeling captures cross-turn dependencies often overlooked in previous work.
(b) \textbf{Intent inference:} By analyzing structural patterns, this taxonomy facilitates the extraction of implicit user intent, offering a deeper understanding of how instructions evolve over multiple turns.
(c) \textbf{Controlled generation:} The taxonomy provides configurable structural parameters that guide task-specific dialogue simulation, allowing for the tailored generation of multi-turn conversations with predefined structural patterns.
This not only enhances dataset diversity but also supports the development of more robust instruction-following models adaptable to varied real-world applications.

We summarize our contributions as follows: 
\vspace{-.1in}
\begin{itemize}
    \setlength{\itemsep}{0pt}
    \setlength{\parskip}{0pt}
    \item \textbf{Structural Flow Taxonomy}: We propose a six-category structural taxonomy for multi-turn instruction-following evaluation, offering an interpretable framework for analyzing dialogue structural flow.
    \item \textbf{StructFlowBench}: We introduce StructFlowBench, a structurally annotated multi-turn benchmark that leverages a structure-driven generation paradigm to better simulate complex dialogue scenarios.
    \item \textbf{Comprehensive LLM evaluation}: We systematically evaluate 13 state-of-the-art LLMs (3 closed-source and 10 open-source), unveiling disparities in structural processing capabilities and providing empirical insights for optimizing dialogue systems.
\end{itemize}

\section{Related Work}

\subsection{Benchmarks for Multi-Turn Dialogues}
The evolution of dialogue evaluation paradigms has progressed from single-turn assessments to sophisticated multi-turn interaction analysis~\cite{wang2023mint,sun-etal-2024-parrot,duan-etal-2024-botchat}. 
Among these, MT-Bench~\cite{zheng2023judging} pioneered this transition by providing methodologies specifically designed to assess a model's ability to handle multi-turn interactions. 
Building upon this, MT-Bench-101~\cite{bai-etal-2024-mt} introduces a more granular evaluation framework to assess fine-grained capabilities. 
Multi-IF~\cite{he2024multi} expands single-turn dialogues into multi-turn interactions by following simple, predefined linear paths.
However, most existing work on multi-turn dialogue evaluation does not prioritize instruction following assessment and overlooks the role of structural information.
MT-Eval~\cite{kwan-etal-2024-mt} explores four types of multi-turn dialogue structures---recollection, expansion, refinement, and follow-up---which partially inspire our structural framework.
However, MT-Eval does not establish a systematic structural framework and lacks integration of various structural aspects for a comprehensive evaluation.

\subsection{Benchmarks for Instruction Following}
Recent instruction following evaluation predominantly employs constraint-based frameworks~\cite{jiang-etal-2024-followbench,zhang2024cfbench,he2024can,zhou2023instruction}.
Constraints are commonly used in these works to guide and evaluate LLM outputs, serving as criteria for assessing instruction-following behavior. 
In our work, we further extend this usage by employing constraints not only for evaluation but also as guidance for model generation.
InfoBench~\cite{qin2024infobench} introduces the Decomposed Requirements Following Ratio (DRFR) metric, which provides a more granular scoring system by breaking down the evaluation of complex instructions into assessments of their individual simple constraints. 
Furthermore, ComplexBench~\cite{wen2024benchmarking} explores instruction-following capabilities in single-turn complex dialogues through empirical studies of constraint composition.
However, prior work on instruction-following evaluation has primarily focused on single-turn interactions, which are less representative of real-world multi-turn usage. 
While some studies have attempted to split complex single-turn instructions into multi-turn dialogues, these approaches do not fully capture the intentionality and goal-oriented nature of users in real-world contexts.
\section{StructFlowBench}

This section introduces the structural flow framework and constraint categories, details the data construction pipeline and benchmark statistics, and outlines the evaluation protocol.

\begin{figure*}[t!]
    \captionsetup{skip=0pt}
	\centering
	\includegraphics[width=\textwidth]{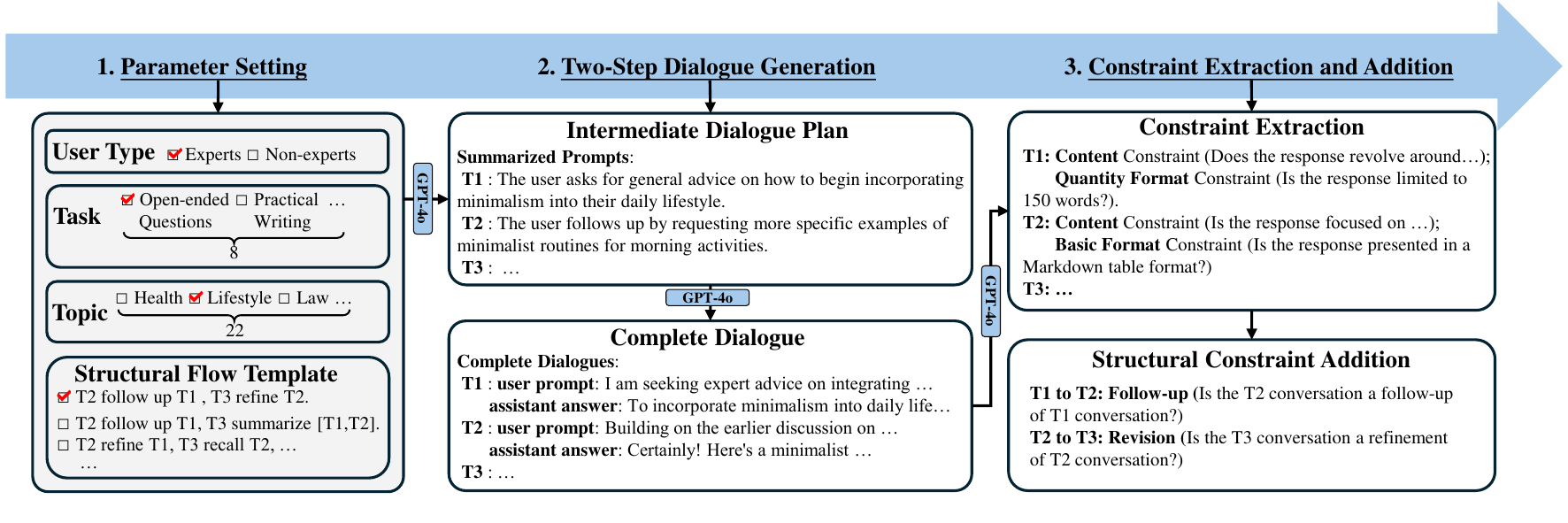}
	\caption{The construction pipeline of \textbf{StructFlowBench}. First, tasks, topics, user types, and structural flow templates are defined. Then, dialogue data is generated in two steps: intermediate dialogue plans (i.e., the summarized prompts) are created from the structural flow, followed by generating complete dialogues from these plans. Finally, intra-turn constraints are extracted by GPT-4o, and structural constraints are added based on the structural flow information.}
	\label{fig:construction_pipeline}
\end{figure*}

\subsection{Structural Flow Taxonomy}\label{sec:structural_flow_taxonomy}

By analyzing existing LLM and real human multi-turn dialogue datasets (such as \textit{WILDCHAT}~\cite{zhao2024wildchat} and \textit{LMSYS-Chat-1M dataset}~\cite{zheng2023lmsys}), we identified and categorized six structural patterns of multi-turn dialogues to enhance the understanding and analysis of conversational structural flow.
Descriptions of these six structures are illustrated in Figure~\ref{fig:intro}.

\textbf{Follow-up:} An adjacent-turn structure where the user's next prompt builds on the previous turn, incorporating details from either the user's previous prompt or the AI's previous response.
This is the most common structure in multi-turn dialogues, typically reflecting the user's intent to explore the topic more deeply.

\textbf{Refinement:} An adjacent-turn structure in which the user modifies or clarifies their immediate previous prompt to improve the AI's response. 
This structure usually reflects dissatisfaction with the prior response and prompts the user to revise the prompt to better convey their concerns.

\textbf{Recall:} A long-range structure where the user refers back to content from two or more turns ago, either to provide context for the current prompt (long-range follow-up) or to seek clarification (long-range refinement).

\textbf{Expansion:} A multi-turn ``fan-out'' structure where the user introduces a main theme and explores related subtopics in subsequent turns.
This structure suggests that the user's following turns are focused on specific subtopics derived from a particular point in the conversation.

\textbf{Summary:} A multi-turn ``fan-in'' structure in which the user requests a consolidation of content from multiple previous turns into a cohesive overview.
This structure acts as the counterpart to expansion, reflecting the need to summarize and condense the information discussed in earlier turns.

\textbf{Unrelatedness:} A conversational structure in which the user's prompt is entirely independent of the previous turn, with no reference to prior content or context.
This structure often occurs in everyday use of LLMs by non-experts, where a new topic is introduced within a previously unrelated dialogue, rather than starting a new conversation.

These six dialogue structures form the Structural Flow Taxonomy, which we use to analyze multi-turn dialogues and construct corresponding structural flows.

\subsection{Constraint Categories}

We categorize our constraints into \textbf{intra-turn constraints} and \textbf{multi-turn structural constraints}.
The definitions and examples of all constraints are provided in Appendix Table~\ref{tab:constraint-definition}, and their distribution is presented in Appendix Table~\ref{tab:constraint-distribution}.

For \textbf{intra-turn constraints}, we synthesize and refine constraint classification systems from several works in this field (e.g., IF-Eval~\cite{zhou2023instruction}, CFBench~\cite{zhang2024cfbench}, FollowBench~\cite{jiang-etal-2024-followbench}). Based on this synthesis, we categorize constraints into eight types: \textit{Inverse Constraint, Style Constraint, Situation Constraint, Keyword/Element Constraint, Basic Format Constraint, Quantity Format Constraint, Template Format Constraint and Content Constraint}.

For \textbf{multi-turn structural constraints}, we define five types of structural constraints, excluding the ``unrelatedness'' structure. 
These constraints are specifically designed to maintain logical coherence and continuity across multiple turns in a dialogue. 
They ensure that the structural relationships between turns are consistent and contextually relevant, enabling a smooth flow of conversation. 
The five types of constraints are aimed at handling key aspects such as follow-ups, refinements, recalls, expansions, and summaries, ensuring that each turn in the dialogue properly connects to the previous ones while adhering to the intended conversational structure.

\subsection{Data Construction Pipeline}

The construction pipeline of StructFlowBench, as shown in Figure~\ref{fig:construction_pipeline}, comprises three main components: parameter setting, two-step dialogue generation, and constraint extraction and addition.
All prompt templates used in the data construction process are included in Appendix~\ref{sec:prompt}, and a sample data instance is provided in Appendix Table~\ref{tab:data-case}.

\subsubsection*{Parameter Setting}

Before dialogue generation, we select parameters such as topic, task, user characteristics, and structural flow template, ensuring comprehensive coverage of the evaluation scope for multi-turn dialogue generation.
For task types, we refer to the taxonomy of ComplexBench~\cite{wen2024benchmarking}, adapting it to our evaluation framework and selecting eight task types.
For topics, we draw from the MT-Bench-101~\cite{bai-etal-2024-mt} framework, making necessary adjustments to suit our context, and ultimately select 22 topics.
For user characteristics, we consider the significant differences in questioning styles and language between experts and non-experts.
Regarding the structural flow templates, we manually designed 14 templates from real dialogue data (WildChat), covering most pairwise combinations of six structural relations.

\begin{table*}[h]
    \begingroup
    \renewcommand{\arraystretch}{0.9}
    \captionsetup{skip=3pt}
    \small
    
    \centering
    \resizebox{\textwidth}{!}{
        \begin{tabular}{lc>{\centering\arraybackslash}p{1.5cm}>{\centering\arraybackslash}p{2cm}>{\centering\arraybackslash}p{2.2cm}>{\centering\arraybackslash}p{2.2cm}>{\centering\arraybackslash}p{2.2cm}}
            \toprule
            \multirow{2}{*}{\textbf{Benchmark}} & \multirow{2}{*}{\textbf{\#Dialogues}} & \textbf{Avg. \#Turns} & \textbf{\#Constraint Types} & \textbf{Fine-grained Constraint} & \textbf{Multi-turn Assessment} & \textbf{Structural Information} \\
            \midrule
            IFEval    & 541  & 1  & 4  & \cmark & \xmark & \xmark \\
            CELLO      & 523  & 1  & 4  & \cmark & \xmark & \xmark \\
            FollowBench    & 820  & 1  & 6  & \cmark & \xmark & \xmark \\
            InfoBench       & 500  & 1  & 5  & \cmark & \xmark & \xmark \\
            CFBench         & 1000  & 1  & 10  & \cmark & \xmark & \xmark \\
            ComplexBench    & 1150  & 1  & 19  & \cmark & \xmark & \xmark \\
            MT-Bench-101    & 1388  & 3.03  & -  & \xmark & \cmark & \xmark \\
            Multi-if    & 4501  & 3  & 24  & \cmark & \cmark & \xmark \\
            MT-Eval     & 168  & 6.96  & -  & \xmark & \cmark &  $\triangle$  \\
            \midrule
            \textbf{StructFlowBench} & 155  & 4.14  & 13  & \cmark & \cmark & \cmark \\
            \bottomrule
        \end{tabular}
    }
    \caption{Comparisons of \textbf{StructFlowbench} with existing benchmarks in terms of constraint types, multi-turn coverage, and structural information. $\triangle$ represents partially satisfied.}
    \label{tab:benchmark-comparison}
    \endgroup
\end{table*}

\subsubsection*{Two-Step Dialogue Generation}
We employ a two-step process to generate a dialogue for a parameter setting. 
The first step uses the structural flow template to generate an intermediate dialogue plan (i.e., summarized prompts) via GPT-4o. 
The detailed prompt template is provided in Appendix Figure~\ref{fig:imtermediate-prompt}.
Locally deployed mini-models perform initial screening and manual inspection of error data to ensure the dialogue plan aligns with the structural flow. 
In the second step, each intermediate dialogue plan is used to generate a complete dialogue, including user prompts and LLM responses via GPT-4o.
The detailed prompt template is provided in Appendix Figure~\ref{fig:complete-prompt}.
This approach ensures high-quality generation of both dialogue content and structure while minimizing manual effort.

\subsubsection*{Constraint Extraction and Addition}
For the complete multi-turn dialogue data, we extract intra-turn constraints using the GPT-4o, followed by manual validation to ensure accuracy. 
Further details are provided in Appendix Figure~\ref{fig:extraction-prompt}.
Based on the structural flow information, we then assign the corresponding multi-turn structural constraints to each dialogue turn.

\subsection{Benchmark Dataset Statistics}

Table~\ref{tab:benchmark-comparison} presents a comparison of related benchmark datasets, evaluating them from three perspectives: fine-grained constraints, multi-turn dialogue assessment, and structural information.
Our StructFlowBench encompasses 8 task types, 22 topics, and 13 constraint types. 
It ultimately includes 155 multi-turn dialogues, comprising a total of 643 turns and 1,775 constraints.
Detailed statistics for tasks and topics are provided in the Appendix~\ref{sec:topic-task}.

\subsection{Evaluation}

\subsubsection*{Evaluation Criteria}
Drawing on the methodology of MT-Bench-101~\cite{bai-etal-2024-mt}, we implemented the ``Golden Context'' approach in our evaluation framework. 
Instead of relying on model-generated contexts, this method uses carefully curated datasets as dialogue histories. 
By providing accurate and consistent contexts for each dialogue turn, it minimizes biases and noise, improving the reliability, fairness, and comparability of response quality assessments across different models.

To achieve a fine-grained evaluation of multi-turn user instructions, we integrate insights from prior studies~\cite{qin2024infobench,wen2024benchmarking,zhang2024cfbench,he2024can} and propose an assessment method based on constraint decomposition and binary question formulation. 
Specifically, we decompose each user instruction into multiple independent constraints and design concise binary questions for each, answered with a simple ``Yes'' or ``No'' to assess satisfaction. 
These binary questions are then aggregated into a checklist that comprehensively covers all critical constraints of the instruction.

Building on this foundation, we further adopt the approach of leveraging state-of-the-art LLMs for evaluation, as outlined in MT-Bench~\cite{zheng2023judging}. 
In our implementation, we use the advanced GPT-4o as the LLM evaluator. 
By providing the evaluator with the golden context, response of the test model, the constraint checklist, and a carefully crafted prompt template, we ensure high consistency and reliability in the evaluation process. 
The prompt template is designed to emphasize key evaluation points, effectively enhancing the accuracy and credibility of the results.

\subsubsection*{Evaluation Metrics}
We adopted several existing metrics, including Constraint Satisfaction Rate (CSR) and Instruction Satisfaction Rate (ISR) ~\cite{zhang2024cfbench}, as well as Decomposed Requirements Following Ratio (DRFR) ~\cite{qin2024infobench}.

The \textbf{Constraint Satisfaction Rate (CSR)} evaluates the average proportion of satisfied constraints across all instructions, calculated as $ \text{CSR} = \frac{1}{m} \sum_{i=1}^{m} \left( \frac{1}{n_i} \sum_{j=1}^{n_i} s_i^j \right) $, where $ m $ represents the total number of instructions, $ n_i $ denotes the number of constraints in the $ i $-th instruction, and $ s_i^j \in \{0, 1\} $ indicates whether the $ j $-th constraint in the $ i $-th instruction is satisfied. 

The \textbf{Instruction Satisfaction Rate (ISR)} measures the proportion of instructions where all constraints are fully satisfied, computed as $ \text{ISR} = \frac{1}{m} \sum_{i=1}^{m} s_i $, where $ s_i \in \{0, 1\} $ indicates whether all constraints in the $ i $-th instruction are satisfied.

The \textbf{Decomposed Requirements Following Ratio (DRFR)} evaluates the overall satisfaction of requirements across all instructions, defined as $ \text{DRFR} = \frac{\sum_{i,j} r'_{i,j}}{\sum_i m_i} $, where $ m_i $ is the number of scoring questions for the $ i $-th instruction, and $ r'_{i,j} $ denotes the result of the $ j $-th scoring question in the $ i $-th instruction.

Despite their utility, these existing metrics have limitations. 
For instance, CSR treats all constraints equally without considering their relative importance, while ISR provides a binary evaluation that may overlook partial fulfillment of constraints. 
To overcome these limitations, we introduce the \textbf{Weighted Constraint Satisfaction Rate (WCSR)}, defined as \textbf{$ \text{WCSR} = \frac{\sum_{j=1}^{n} w_j \cdot s_j}{\sum_{j=1}^{n} w_j}, $} which incorporates weighted factors to account for the varying significance of different constraint types.
Here, $ n $ denotes the total number of constraints, $ w_j $ represents the weight assigned to the $ j $-th constraint, and $ s_j \in \{0, 1\} $ indicates whether the $ j $-th constraint is satisfied. 
In our framework, intra-turn constraints are assigned a weight of $ w_r = 1 $, whereas structural constraints, which play a critical role in ensuring coherence and correctness, are given a higher weight of $ w_s = 2 $.

The introduction of WCSR provides a more nuanced evaluation by emphasizing important constraints through weighted assessments. 
This improves the precision and relevance of evaluations, enhancing the reliability of LLMs in meeting complex requirements.
\begin{table*}[htbp]
    \begingroup
    \renewcommand{\arraystretch}{0.9}
    \captionsetup{skip=3pt}

    \centering
    \resizebox{\textwidth}{!}{
        \begin{tabular}{lccccc|cccc}
            \toprule
            \textbf{Model Name} & \textbf{follow-up} & \textbf{refinement} & \textbf{expansion} & \textbf{summary} & \textbf{recall} & \textbf{CSR} & \textbf{ISR} & \textbf{WCSR} & \textbf{DRFR} \\
            \midrule
            Deepseek-v3 & \textbf{\underline{0.99}}& \textbf{\underline{0.8}}& \textbf{\underline{0.92}}& \textbf{\underline{1.0}} & \textbf{\underline{1.0}} & \textbf{\underline{0.97}}& \textbf{\underline{0.93}}& \textbf{\underline{0.96}} & \textbf{\underline{0.98}} \\
            Gemini-1.5-Pro & 0.97& 0.78& 0.91& \textbf{\underline{1.0}} & 0.94& 0.96 & 0.91& 0.95 & 0.96 \\
            GPT-4o & 0.98& 0.78& 0.88& 0.97& 0.91& 0.96& 0.9 & 0.95& 0.96 \\
            Claude-3.5-Sonnet & 0.98 & \textbf{\underline{0.8}}& 0.88& \textbf{\underline{1.0}} & 0.91& 0.95 & 0.89& 0.94 & 0.95\\
            GLM-4-9B-Chat & 0.95 & 0.75& 0.84& 0.97& 0.94& 0.95 & 0.87 & 0.93 & 0.95 \\
            Qwen2.5-14B-Instruct & 0.97& 0.73& 0.87& 0.97& 0.97& 0.93 & 0.84& 0.92 & 0.93 \\
            Qwen2.5-7B-Instruct & 0.95& 0.76& 0.9& 0.94& 0.97& 0.93 & 0.84& 0.92 & 0.93 \\
            Deepseek-R1-Distill-Qwen-7B & 0.91& 0.62& 0.85& 0.86& 0.78 & 0.81 & 0.7& 0.8 & 0.82 \\
            DeepSeek-R1-Distill-Llama-8B & 0.94& 0.73& 0.82& 0.89& 0.84 & 0.87& 0.79& 0.86& 0.87\\
            Llama-3.1-Instruct-8B & 0.96& 0.71& 0.84& 0.79& 0.94& 0.84 & 0.69& 0.83 & 0.85 \\
            Phi-3.5-mini-instruct & 0.94& 0.68& 0.87& 0.94& 0.94& 0.88 & 0.74& 0.87 & 0.87 \\
            Yi-6B-Chat & 0.98 & 0.62& 0.87 & 0.84& 0.94& 0.86 & 0.7& 0.84& 0.86 \\
            Mistral-7B-Instruct-v0.3 & 0.97& 0.59& 0.87& 0.71& 0.97& 0.76 & 0.57& 0.76 & 0.77 \\
            \bottomrule
        \end{tabular}
    }
    \caption{\textbf{StructFlowBench} rated by \textbf{GPT-4o}. The left side of the figure displays the performance of various models on the five basic structural constraints, with \textbf{accuracy} used as the evaluation metric, while the right side presents their performance on the four key metrics.}
    \label{tab:main-results}
    \endgroup
\end{table*}

\section{Experiments}

\subsection{Experimental Setup}
We evaluate 13 popular LLMs on StructFlowBench, including 3 closed-source models (GPT-4o~\cite{hurst2024gpt}, Claude-3.5-Sonnet~\cite{anthropic2024claude} and Gemini-1.5-Pro~\cite{team2024gemini} ) and 10 open-source models: Llama-3.1-Instruct-8B~\cite{dubey2024llama}, Mistral-7B-Instruct-v0.3~\cite{jiang2023mistral}, Qwen2.5-7B-Instruct, Qwen2.5-14B-Instruct~\cite{yang2024qwen2}, Yi-6B-Chat~\cite{young2024yi}, Phi-3.5-mini-instruct~\cite{abdin2024phi}, GLM-4-9B-Chat~\cite{glm2024chatglm}, Deepseek-R1-Distill-Llama-8B, Deepseek-R1-Distill-Qwen-7B~\cite{guo2025deepseek} and DeepSeek-v3~\cite{liu2024deepseek}.
More details on these evaluated models can be found in Appendix Table~\ref{tab:model-link}.

\subsection{Main Results}
\subsubsection*{Overall Results}

Table~\ref{tab:main-results} presents a comprehensive evaluation of 13 representative LLMs on StructFlowBench, covering four key metrics as well as assessments of structural constraints. 
The detailed results, categorized by intra-turn constraints and task types, are provided in the Appendix~\ref{sec:detailed results}.

The recently released DeepSeek-v3 outperforms all other models across all metrics, demonstrating its exceptional capability in fine-grained constraint satisfaction and multi-turn dialogue structure understanding. 
Gemini-1.5-Pro and GPT-4o closely follow, achieving comparable performance in intra-turn constraints but showing slightly weaker results in adhering to structural constraints for multi-turn dialogues. 
Claude-3.5-Sonnet, GLM-4-9B-Chat, Qwen2.5-14B-Instruct, and Qwen2.5-7B-Instruct also exhibit strong instruction-following capabilities, with CSR exceeding 94\%. 
Notably, all seven of these models achieve high DRFR scores, indicating their strong ability to follow fine-grained instructions.

In contrast, mid-tier models such as Deepseek-R1-Distill-Llama-8B, Llama-3.1-8B-Instruct, Phi-3.5-Mini-Instruct, and Yi-6B-Chat perform reasonably well but exhibit greater instability, particularly in ISR and WCSR. 
While they handle simpler constraints effectively, they struggle with maintaining consistency when processing complex instructions and multi-turn dialogue structures. 
The weakest performers in multi-turn instruction following are Deepseek-R1-Distill-Qwen-7B and Mistral-7B-Instruct-v0.3, revealing significant deficiencies in natural interaction scenarios.

A particularly interesting observation is that Deepseek-R1-Distill-Llama-8B, distilled from Llama-3.1-8B, outperforms Llama-3.1-8B-Instruct across all metrics, demonstrating the effectiveness of the distillation process. 
However, Deepseek-R1-Distill-Qwen-7B, distilled from Qwen2.5-Math-7B, underperforms due to its origin from a model optimized primarily for mathematical reasoning tasks, which inherently makes it weaker in multi-turn dialogue instruction following compared to Qwen2.5-7B-Instruct.

One particularly noteworthy outcome is that DeepSeek-v3, an open-source model, surpasses its closed-source counterparts in multi-turn instruction-following evaluations. 
This result is encouraging for both the research community and the open-source ecosystem, suggesting that the theoretical advancements and training methodologies behind DeepSeek-v3 could offer valuable insights for improving LLMs in multi-turn instruction-following tasks.

\subsubsection*{Structural-Constraint-Categorized Performance}

The evaluated LLMs exhibit strong performance in follow-up structures, with nearly all models excelling in maintaining contextual continuity and generating coherent responses.
Additionally, most models handle recall structures well, demonstrating their ability to reference prior conversational turns effectively.
However, performance varies when dealing with more complex structures such as summary and expansion.
DeepSeek-v3 and proprietary models outperform the others, indicating their superior capability in nuanced content condensation and elaboration.
In contrast, refinement tasks pose a significant challenge across all models.  
Even the strongest model, DeepSeek-v3, achieves only 0.8 in this category, highlighting the inherent difficulty of processing refinements accurately and maintaining coherence when adapting to modified user inputs.
While LLMs exhibit strong instruction-following abilities in structured dialogue, refinement remains the most challenging task, requiring improvements in dynamic response adaptation. 
Future advancements should focus on enhancing models' flexibility in refining responses based on iterative user feedback, ensuring more robust handling of complex multi-turn interactions.

\subsubsection*{Intra-Turn-Constraint-Categorized Performance}

The evaluation of LLMs across various constraint dimensions highlights their strengths and weaknesses in following specific instructions. 
DeepSeek-v3, Gemini-1.5-Pro, and GPT-4o achieve near-perfect satisfaction rates, demonstrating strong capabilities in fine-grained instruction following.
Most other models also perform well in rule-based constraints, such as Inverse Constraint, Keyword/Element Constraint, Style Constraint, and Situation Constraint.
However, performance drops noticeably in format-related constraints, including Basic Format Constraint, Template Format Constraint, and Quantity Format Constraint, indicating that rigid format adherence remains a significant challenge, even for top-performing models.
Overall, while LLMs effectively handle intra-turn constraints, their ability to maintain format consistency remains a key limitation. 
Addressing this challenge requires further advancements in structured output generation and adherence to strict formatting requirements.

\begin{figure*}[htbp]
    \captionsetup{skip=0pt}
	\centering
	\includegraphics[width=\textwidth]{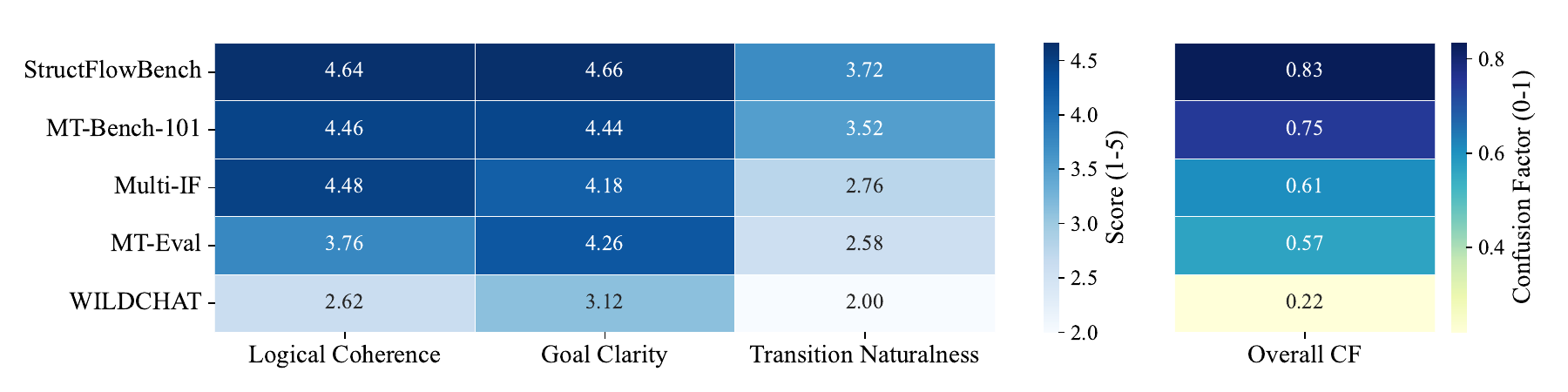}
	\caption{The comprehensive complex scenario evaluation heatmap of five multi-turn dialogue datasets.}
	\label{fig:heatmap}
\end{figure*}

\subsubsection*{Task-Categorized Performance}

We evaluated various models across seven NLP tasks and a mixed task.
Unlike the constraint-categorized evaluation, where DeepSeek-v3 led across all metrics, the task-based analysis presents a more nuanced picture. 
DeepSeek-v3 remains the overall best-performing model but leads only in Fact-based Questions, Professional Writing, Practical Writing, and Casual Chat.
Gemini-1.5-Pro outperforms others in Open-ended Questions and Creative Writing, while Claude-3.5-Sonnet achieves the highest performance in Fact-based Questions and Task-oriented Role-playing. 
Meanwhile, GPT-4o excels in the Mixture task type, reflecting its strength in handling diverse instructions across domains. 
These results highlight the varying strengths of these top-tier models across different tasks.
Following the top four models, GLM-4-9B-Chat, Qwen2.5-14B-Instruct, and Qwen2.5-7B-Instruct maintain consistently strong performance across all tasks. 
Their stability, combined with their significantly smaller parameter sizes compared to the leading models, makes them highly cost-effective alternatives.
In contrast, the remaining models all exhibit noticeable weaknesses in at least one task category, with Mistral-7B-Instruct-v0.3 underperforming across nearly all tasks, revealing a clear performance gap.

\subsection{Further Analysis}

\subsubsection{Complex Scenario Suitability Study}
This study aims to verify whether the multi-turn dialogue dataset we have constructed more closely aligns with real-world complex use cases.
To achieve this, we designed an experiment to analyze three key properties of dialogue: logical coherence, goal clarity, and transition naturalness. 
The datasets used in this experiment include our StructFlowBench, three other multi-turn dialogue evaluation datasets (MT-Bench-101, Multi-if, and MT-Eval), and a real-world dialogue dataset, WILDCHAT.

\textbf{Data Preparation:} For each dataset, we randomly selected 50 English multi-turn dialogue samples, ensuring a diverse representation of dialogue types.

\textbf{Evaluation Protocol:} To quantify how well the dialogues meet complex scenario requirements, we employed GPT-4o for automated scoring. 
Each dialogue was evaluated based on its performance in the following areas:
\vspace{-.1in}
\begin{itemize}
    \setlength{\itemsep}{0pt}
    \setlength{\parskip}{0pt}
    \item \textbf{Logical Coherence:} Evaluates whether the dialogue is logically consistent and free of abrupt or unreasonable shifts.
    \item \textbf{Goal Clarity:} Assesses whether the dialogue clearly communicates the task's goals and ensures both the user's and system's intentions are transparent.
    \item \textbf{Transition Naturalness:} Judges whether transitions between dialogue turns are smooth and natural, without awkward or forced shifts.
\end{itemize}
\vspace{-.1in}
Each property was scored on a scale from 1 to 5, where 1 indicates complete failure to meet the expected standard, and 5 represents perfect alignment with complex scenario requirements.

\textbf{Confusion Factor (CF):}
To further evaluate the datasets, we introduced the Confusion Factor (CF), which quantifies the proportion of dialogues in each dataset that scored 4 or higher, indicating they were mistakenly perceived as real-world interactions. 
The CF is calculated as follows:
$$
\text{CF} = \frac{\text{Number of dialogues with average score} \geq 4}{\text{Total number of dialogues}},
$$
By comparing the CF values of our StructFlowBench dataset with those of others, we can assess whether our dataset outperforms the others in terms of alignment with complex scenarios.

\textbf{Results and Discussion:}
The results are presented as a heatmap, as shown in Figure~\ref{fig:heatmap}.
StructFlowBench achieves the highest scores across all three evaluation dimensions, leading with a confusion factor of 0.83. 
MT-Bench-101, with its comprehensive dialogue generation process and rigorous human proofreading, also produces high-quality dialogues and ranks closely behind with strong scores.
In contrast, the WILDCHAT real multi-turn dialogue dataset, containing one million dialogues, exhibits generally low quality. 
Although we performed preliminary filtering on the WILDCHAT data, such as considering prompt length and dialogue content, the extracted dialogues still failed to meet the ideal quality standards
As a result, WILDCHAT performed the worst across the three evaluation dimensions for data-driven simulated scenarios.

\subsubsection{Fail Case Analysis of Refinement}
\begin{figure}[t!]
    \captionsetup{skip=0pt}
	\centering
	\includegraphics[width=\linewidth]{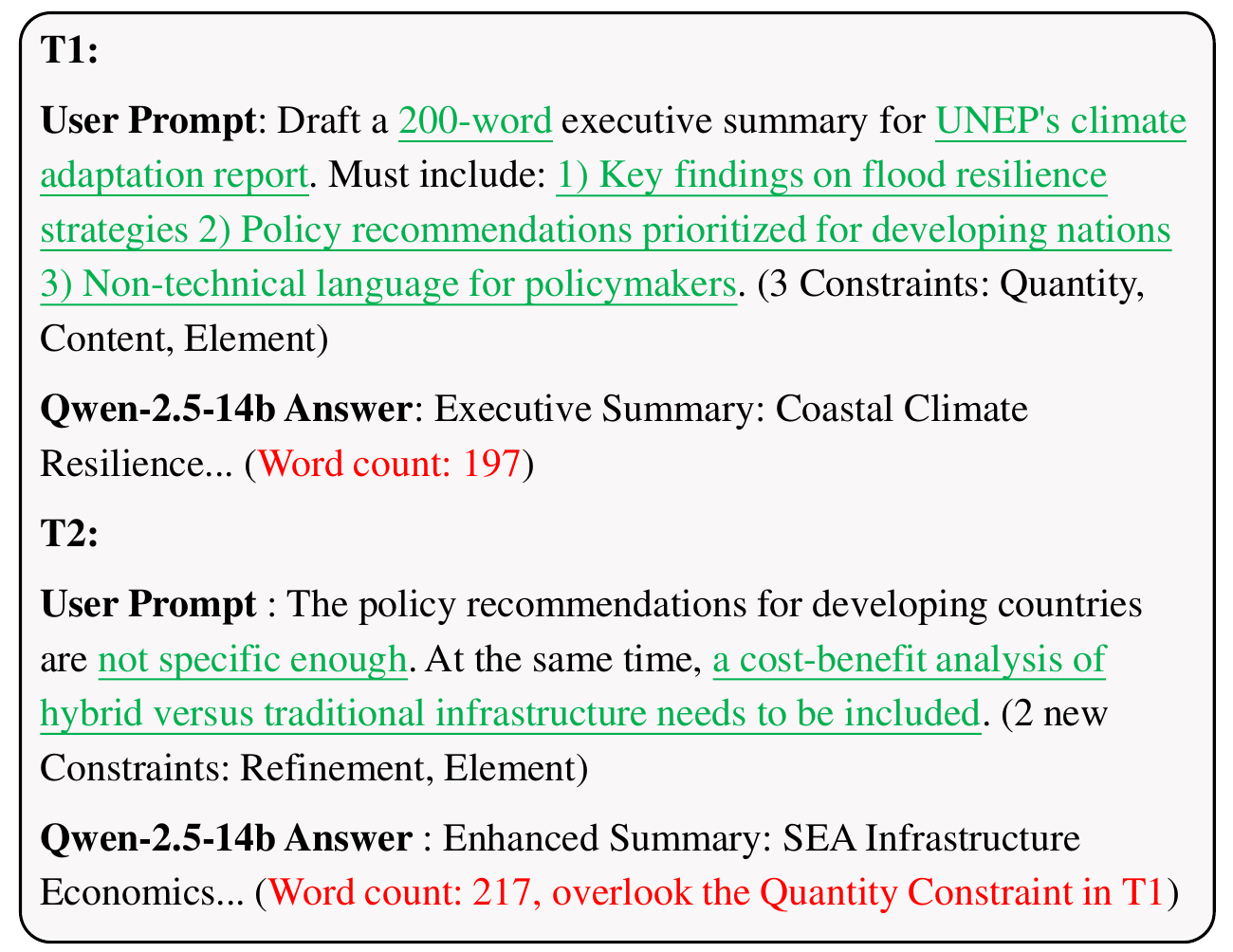}
	\caption{A Fail Case of Refinement.}
	\label{fig:fail-case}
\end{figure}

Given that none of the models achieved optimal performance on the Refinement structure, we conducted an in-depth analysis of failure cases specific to this category. 
Our analysis suggests that the high error rate in Refinement tasks stems from the model’s difficulty in distinguishing refinement from follow-up instructions, particularly in multi-turn settings. 
In refinement scenarios, only specific constraints should be updated, while others—often introduced in earlier turns—must remain intact. 
However, models frequently overlook these earlier constraints, failing to preserve them in the revised responses. 
This forgetting behavior leads to violations of the intended dialogue structure and undermines the overall satisfaction of constraints.
A concrete example of such a failure is illustrated in Figure~\ref{fig:fail-case}.

\subsection{Human Verification}

We extracted 30 dialogues from the output of Qwen2.5-7B-Instruct and invited two domain experts to conduct a comprehensive and detailed evaluation of the results.
The task requirements provided to the human evaluators are shown in Appendix Figure~\ref{fig:task-requirement}.
Specifically, the experts were instructed to assess adherence to constraints through a binary evaluation protocol. 
Each constraint was assigned a ``Yes'' for full alignment and a ``No'' for any violation.
The results showed that the Kappa coefficient between GPT-4o's evaluations and those of the experts was approximately 0.75. 
This indicates that utilizing advanced LLMs, like GPT-4o, to assess the quality of outputs from other models is a reliable approach, effectively reducing both subjective bias and the time costs associated with relying solely on human evaluation.
\section{Conclusion}

In this work, we address key limitations in current multi-turn instruction-following research by introducing StructFlowBench, a novel benchmark designed to capture the structural intricacies of complex dialogue scenarios.
By incorporating a dual-constraint evaluation system and a six-category structural flow taxonomy, we provide a more comprehensive framework for assessing the logical coherence, goal clarity, and transition naturalness of multi-turn dialogues. 
Our evaluations of 13 representative LLMs reveal critical insights into the structural processing capabilities of both closed-source and open-source models, offering valuable guidance for future advancements in instruction-following systems. 
Through StructFlowBench, we lay the foundation for more robust, realistic, and contextually aware dialogue systems.

\section*{Limitations}
The current StructFlowBench is more of an exploration of new directions in evaluating multi-turn dialogue instruction following, rather than a targeted effort to increase the difficulty of the evaluation set. 
As a result, leading models tend to achieve near-maximum scores on many metrics. 
Future work can enhance the difficulty by extending the number of dialogue turns, introducing more constraints per turn, and incorporating more challenging tasks such as complex reasoning and retrieval-augmented generation (RAG).

In addition to increasing task difficulty, another important direction lies in improving the structural design itself.
Currently, the structural flow in StructFlowBench is designed with a single linear relationship to facilitate analysis and data generation. 
For instance, if the third turn dialogue serves as both a recall structure to the first turn and a follow-up structure to the second turn, the current approach retains only the recall relationship while disregarding other structural dependencies. 
This simplification may limit the comprehensive modeling of hierarchical dialogue structures. 
Future work should extend the structural flow framework to simultaneously capture multiple coexisting dialogue relationships, thereby providing a more holistic representation of multi-turn dialogue complexity.

\section*{Ethics Statement}
This study utilizes GPT-4o to generate multi-turn dialogue data and annotate constraints, with manual review to filter out inappropriate content. 
However, unintended biases in GPT-4o's generation process, as well as potential oversight during human review, may result in residual errors or biases in the dataset. 
While we have made every effort to ensure data quality and mitigate these issues, completely eliminating them remains challenging. 
Additionally, since this dataset is publicly available, there is a risk of misuse for model training, which may compromise the validity of our benchmark. 
Therefore, we encourage the research community to exercise caution when using this dataset and to complement it with other evaluation methods to ensure comprehensive and fair model assessment.

\section*{Acknowledgments}
The authors would like to thank the anonymous reviewers for their valuable comments.
This work is supported by the National Key Research and Development Program of China (No.2023YFF0905400), the National Natural Science Foundation of China (No.U2341229) and the Reform Commission Foundation of Jilin Province (No.2024C003).

\bibliography{custom}

\clearpage
\appendix

\section{Details of Topics and Tasks}
\label{sec:topic-task}
\begin{itemize}
    \item \textbf{Topic:} Our dataset is generated across a diverse range of 22 topics, including health, history, science, technology, digital media, automotive, astronomy, geography, lifestyle, literature, physics, finance, stocks, law, humanities, entertainment, music, fashion, art, environment, psychology, and a mixed category that incorporates multiple topics. This broad coverage ensures that our data spans multiple domains, capturing a wide array of fields and areas of interest.
    \item \textbf{Task:} StructFlowBench comprises seven NLP tasks and one mixed-category task, with their exact distribution detailed in Table~\ref{tab:task-distribution}.
\end{itemize}

\begin{table}[htbp]
\centering
\begin{tabular}{l|r}
\toprule
\textbf{Category} & \textbf{\#Dialogues} \\
\midrule
Fact-based Questions & 25 \\
Open-ended Questions & 20 \\
Practical Writing & 26 \\
Creative Writing & 21 \\
Professional Writing & 21 \\
Casual Chat & 15 \\
Task-oriented Role Play & 17 \\
Mixture & 10 \\
\midrule
Total & 155 \\
\bottomrule
\end{tabular}
\caption{Task distribution of \textbf{StructFlowBench} dataset.}
\label{tab:task-distribution}
\end{table}

\begin{table*}[htbp]
    \centering
    \begin{tabular}{p{3.5cm}p{7cm}p{5cm}}
        \toprule
        \textbf{Constraint Name} & \textbf{Definition} & \textbf{Example} \\
        \midrule
        Content Constraint & The response must strictly focus on the specified content scope and avoid any deviation from the topic. & Is the response focused on recommending a graphics card for a gaming PC? \\
        \midrule
        Keyword/Element Constraint & The response must include specific words or elements as required. & Must contain the word “Artificial Intelligence” in your answer. \\
        \midrule
        Style Constraint & The response must be generated in a specific writing style, such as formal, humorous, poetic, etc.  & Please write a report in a formal style. \\
        \midrule
        Basic Format Constraint & The output must adhere to a specified basic format, such as JSON, XML, CSV, Table, Markdown, etc.  & Please output the following data in JSON format. \\
        \midrule
        Quantity Format Constraint & The response must meet a precise requirement for the number of characters, words, sentences, or paragraphs as specified.  & Please provide an answer in no more than 100 words. \\
        \midrule
        Template Format Constraint & The response must follow a predefined template structure, such as starting with a specific phrase, ending with a certain statement, or using a custom template provided by the user.  & Please follow the template below to generate your content. \\
        \midrule
        Situation Constraint & The response must be tailored to a given scenario or perspective, such as responding from a specific identity or context. & Imagine you are an experienced doctor and respond to the following health-related questions.  \\
        \midrule
        Inverse Constraint & The response must deliberately exclude or avoid certain constraints, such as not containing a specific keyword, not involving a particular element, or not using a certain language style.  & Make sure your response does not involve duscrimination or politics. \\
        \midrule
        Follow-up Constraint & The response must reflect a follow-up structure, building upon the previous turn by incorporating elements from the user’s or AI’s prior input to ensure dialogue continuity. & Is the T2 turn a follow-up of T1? \\
        \midrule
        Refinement Constraint & The response must conform to a refinement structure, where the user modifies or clarifies their immediately preceding prompt to improve relevance or accuracy while maintaining context. & Is the T3 turn a refinement of T2? \\
        \midrule
        Expansion Constraint & The response must align with an expansion structure, where a main theme is introduced and related subtopics are explored across multiple turns with thematic continuity. & Is the [T2,T3] turns an expansion of the T1? \\
        \midrule
        Summary Constraint & The response must follow a summary structure, consolidating information from multiple previous turns into a coherent and concise overview with clarity and completeness. & Is the T4 turn a summary of [T1,T2,T3]  \\
        \midrule
        Recall Constraint & The response must adhere to a recall structure, referencing content from two or more turns earlier to re-establish context or seek clarification, ensuring long-range coherence. & Is the T5 turn a recall of T2? \\
        \bottomrule
    \end{tabular}
    \caption{Constraint System of \textbf{StructFlowBench}}
    \label{tab:constraint-definition}
\end{table*}

% \section{Details of Constraints}
% \label{sec:constraint}
% The definitions and examples of all constraints are provided in Table~\ref{tab:constraint-definition}, and their distribution is presented in Table~\ref{tab:constraint-distribution}.

\begin{table*}[htbp]
    \centering
    \resizebox{\textwidth}{!}{
        \begin{tabular}{ccccc|cccccccc}
            \toprule
            \textbf{Follow-up} & \textbf{Refinement} & \textbf{Expansion} & \textbf{Summary} & \textbf{Recall} & \textbf{C1} & \textbf{C2} & \textbf{C3} & \textbf{C4} & \textbf{C5} & \textbf{C6} & \textbf{C7} & \textbf{C8}\\
            \midrule
            95 & 32 & 156 & 63 & 118 & 505 & 153 & 140 & 105 & 175 & 98 & 83 & 52 \\
            \bottomrule
        \end{tabular}
    }
    \caption{The constraints distribution of \textbf{StructFlowBench}. \textit{Follow-up}, \textit{Refinement}, \textit{Expansion}, \textit{Summary}, \textit{Recall} denote the structural constraints. The designations C1 - C8 denote the Constraint types of \textit{Content Constraint, Keyword/Element Constraint, Style Constraint, Basic Format Constraint, Quantity Format Constraint, Template Format Constraint, Situation Constraint, Inverse Constraint}}
    \label{tab:constraint-distribution}
\end{table*}

\section{Detailed Results Categorized by Intra-turn Constraints and Task Types}
\label{sec:detailed results}
\begin{figure*}[htbp]
	\centering
	\includegraphics[width=\textwidth]{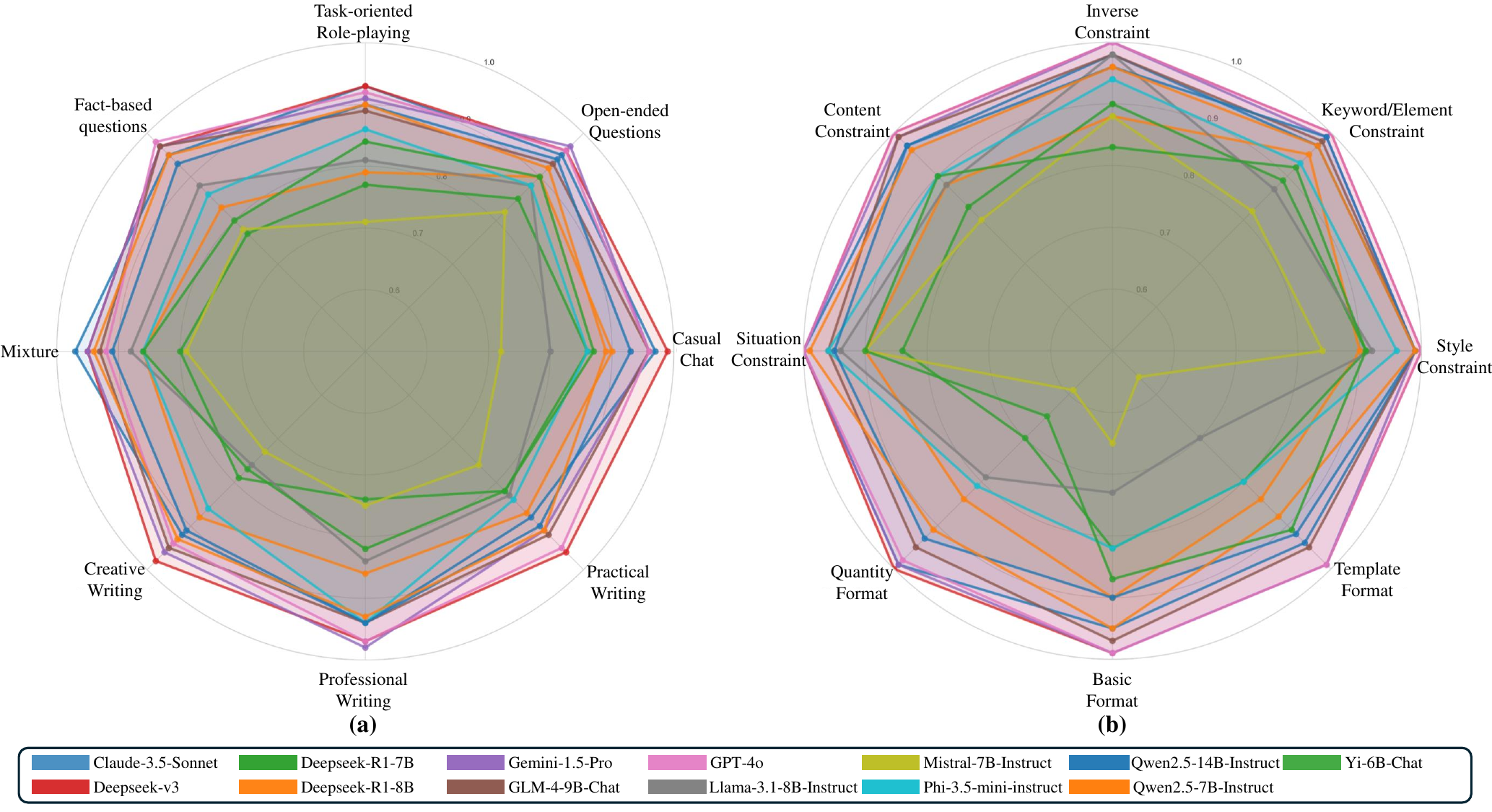}
	\caption{The radar chart of intra-turn-constraint-categorized performance (a) and task-categorized performance (b).}
	\label{fig:radar}
\end{figure*}

Table~\ref{tab:intra-turn-results} presents the intra-turn constraints performance of various models on StructFlowBench, while Table~\ref{tab:task-results} illustrates the task-categorized performance. 
Additionally, Figure~\ref{fig:radar} provides a radar chart comparing both perspectives.

\begin{table*}[htbp]
    \centering
    \resizebox{\textwidth}{!}{
        \begin{tabular}{l>{\centering\arraybackslash}p{2.5cm}>{\centering\arraybackslash}p{2.5cm}>{\centering\arraybackslash}p{2.5cm}>{\centering\arraybackslash}p{2.5cm}>{\centering\arraybackslash}p{3.5cm}>{\centering\arraybackslash}p{3cm}>{\centering\arraybackslash}p{3cm}>{\centering\arraybackslash}p{2.5cm}}
            \toprule
            \multirow{2}{*}{\textbf{Model Name}} & \textbf{Inverse Constraint} & \textbf{Keyword/Element Constraint} & \textbf{Style Constraint} & \textbf{Situation Constraint} & \textbf{Basic Format Constraint} & \textbf{Quantity Format Constraint} & \textbf{Template Format Constraint} & \textbf{Content Constraint} \\
            \midrule
            Deepseek-v3 & \textbf{\underline{1.0}}& \textbf{\underline{1.0}}& \textbf{\underline{1.0}}& \textbf{\underline{1.0}} & \textbf{\underline{0.99}} & \textbf{\underline{1.0}} & \textbf{\underline{0.99}}& \textbf{\underline{1.0}} \\
            Gemini-1.5-Pro & \textbf{\underline{1.0}}& 0.99& 0.99& \textbf{\underline{1.0}} & \textbf{\underline{0.99}}& 0.99 & \textbf{\underline{0.99}}& 0.99 \\
            GPT-4o & \textbf{\underline{1.0}} & \textbf{\underline{1.0}}& \textbf{\underline{1.0}}& \textbf{\underline{1.0}}& \textbf{\underline{0.99}}& 0.98& \textbf{\underline{0.99}} & \textbf{\underline{1.0}}\\
            Claude-3.5-Sonnet & 0.98 & 0.97& 0.99& \textbf{\underline{1.0}} & 0.95& 0.99 & 0.94& 0.97\\
            GLM-4-9B-Chat & 0.98 & 0.98& 0.99& 0.96& 0.97& 0.95 & 0.95 & 0.99 \\
            Qwen2.5-14B-Instruct & 0.96& 0.99& 0.99& 0.95& 0.9& 0.93 & 0.92& 0.97 \\
            Qwen2.5-7B-Instruct & 0.96& 0.97& 0.99& 0.99& 0.95& 0.91 & 0.88& 0.96 \\
            Deepseek-R1-Distill-Qwen-7B & 0.9& 0.89& 0.91& 0.84& 0.82& 0.7 & 0.8 & 0.83 \\
            DeepSeek-R1-Distill-Llama-8B & 0.88& 0.95& 0.9& 0.9& 0.9& 0.84& 0.84& 0.88\\
            Llama-3.1-Instruct-8B & 0.98& 0.87& 0.92& 0.94& 0.73& 0.79 & 0.7& 0.88\\
            Phi-3.5-mini-instruct & 0.94& 0.93& 0.96& 0.96& 0.82& 0.81 & 0.8& 0.9\\
            Yi-6B-Chat & 0.83 & 0.92& 0.91 & 0.9& 0.87& 0.65 & 0.91& 0.9 \\
            Mistral-7B-Instruct-v0.3 & 0.88& 0.82& 0.84& 0.9& 0.65& 0.59 & 0.56 & 0.8 \\
            \bottomrule
        \end{tabular}
    }
    \caption{The intra-turn constraints performance of various models on \textbf{StructFlowBench}.}
    \label{tab:intra-turn-results}
\end{table*}

\begin{table*}[htbp]
    \centering
    \resizebox{\textwidth}{!}{
        \begin{tabular}{l>{\centering\arraybackslash}p{2.5cm}>{\centering\arraybackslash}p{2.5cm}>{\centering\arraybackslash}p{2.5cm}>{\centering\arraybackslash}p{2.5cm}>{\centering\arraybackslash}p{2.5cm}c>{\centering\arraybackslash}p{3cm}c}
            \toprule
            \multirow{2}{*}{\textbf{Model Name}} & \textbf{Fact-based Questions} & \textbf{Open-ended Questions} & \textbf{Professional Writing} & \textbf{Practical Writing} & \textbf{Creative Writing} & \multirow{2}{*}{\textbf{Casual Chat}} & \textbf{Task-oriented Role-playing} & \multirow{2}{*}{\textbf{Mixture}} \\
            \midrule
            Deepseek-v3 & \textbf{\underline{0.93}}& 0.96& \textbf{\underline{0.99}}& \textbf{\underline{0.96}} & 0.97 & \textbf{\underline{0.98}} & 0.95& 0.97 \\
            Gemini-1.5-Pro & 0.91& \textbf{\underline{0.97}}& 0.96& 0.91 & \textbf{\underline{0.98}}& 0.96 & 0.95& 0.97 \\
            GPT-4o & 0.92 & 0.96& 0.96& 0.95& 0.97& 0.94& 0.92 & \textbf{\underline{0.98}}\\
            Claude-3.5-Sonnet & \textbf{\underline{0.93}} & 0.95& 0.97& 0.88 & 0.94& 0.92 & \textbf{\underline{0.97}}& 0.95\\
            GLM-4-9B-Chat & 0.89 & 0.93& 0.96& 0.92& 0.94& 0.95 & 0.93 & 0.97 \\
            Qwen2.5-14B-Instruct & 0.9& 0.94& 0.93& 0.9& 0.94& 0.91 & 0.91& 0.93 \\
            Qwen2.5-7B-Instruct & 0.9& 0.92& 0.89& 0.91& 0.93& 0.93 & 0.94& 0.95 \\
            Deepseek-R1-Distill-Qwen-7B & 0.77& 0.85& 0.86& 0.82& 0.74& 0.79 & 0.8 & 0.77 \\
            DeepSeek-R1-Distill-Llama-8B & 0.79& 0.9& 0.9& 0.87& 0.86& 0.88& 0.86& 0.83\\
            Llama-3.1-Instruct-8B & 0.81& 0.88& 0.8& 0.83& 0.84& 0.76 & 0.88& 0.88\\
            Phi-3.5-mini-instruct & 0.86& 0.88& 0.86& 0.84& 0.94& 0.86 & 0.86& 0.86\\
            Yi-6B-Chat & 0.84 & 0.9& 0.87 & 0.82& 0.82& 0.77 & 0.86& 0.8 \\
            Mistral-7B-Instruct-v0.3 & 0.71& 0.82& 0.72& 0.76& 0.75& 0.73 & 0.79 & 0.78 \\
            \bottomrule
        \end{tabular}
    }
    \caption{Task-categorized performance of various models on \textbf{StructFlowBench}.}
    \label{tab:task-results}
\end{table*}

% \section{Case of Data}
% \label{sec:case}
% Table~\ref{tab:data-case} presents a sample case from StructFlowBench.

\begin{table*}[htbp]
    \centering
    \begin{tabular}{p{0.2\linewidth} p{0.8\linewidth}}
        \toprule
        \textbf{User purpose} 
        & The user aims to develop a financial plan for a fictional character by interacting with the assistant as a financial advisor.The user wants to learn about different music genres and styles to enhance their personal music knowledge and broaden their music listening experience. \\
        \midrule
        \textbf{Structure} 
        & "source": "T1","target": "T2","relation": "follow-up" \\
        & "source": "T1","target": "T3","relation": "recall" \\
        & "source": "T3","target": "T4","relation": "unrelatedness" \\
        & "source": "T4","target": "T5","relation": "refinement" \\
        \midrule
        \textbf{Summarized Prompts} 
        & "T1" : "The user asks the assistant, role-playing as a financial advisor, to provide a general strategy for a young professional who wants to start saving for retirement." \\
        & ... \\
        & "T5": "The user modify the detail level in last round's prompt to request a deeper dive into the unique instruments used in each genre for better understanding of their sounds." \\
        \midrule
        \textbf{Complete Dialogue}
        & "name": "T1", \\
        & "user prompt": "Imagine I am a young professional entering the workforce. As my financial advisor, could you...", \\
        & "assistant answer": "Certainly! Here's a comprehensive strategy for..." \\
        &...\\
        & "name": "T5", \\
        & "user prompt": "In order to delve deeper into the musical intricacies ... Please format the response as a table and ..." \\
        & "assistant answer": "Certainly! Here is a detailed examination of the unique instruments associated with each genre in a table format:..." \\
        \midrule
        \textbf{Check Lists} 
        & "name":"T1" \\
        & "Situation Constraint":"Is the response given from the perspective of a financial advisor?" \\
        & "Keyword/Element Constraint":"Does the response include specific keywords such as... ?" \\
        & ... \\
        & "name":"T5" \\
        & "Basic Format Constraint":"Is the response formatted as a table?" \\
        & "Refinement Constraint":"Is the T5 conversation a refinement of T4 conversation?" \\
        \bottomrule
    \end{tabular}
    \caption{An example of synthetic data.}
    \label{tab:data-case}
\end{table*}

% \section{Details of Models}
% \label{sec:model-link}
% All the details about the evaluated models are provided in Table~\ref{tab:model-link}.

\begin{table*}[htbp]
    \centering
    \begin{tabular}{l p{4cm} p{2cm} p{8cm}}
        \toprule
        \textbf{Model} & & \textbf{Paremeters (Billions)} & \textbf{Model Link} \\
        \midrule
        GPT & GPT-4o & $\sim 200$ & \href{https://platform.openai.com/docs/models#gpt-4o}{https://platform.openai.com/docs/models\#gpt-4o} \\
        \midrule
        Claude & Claude-3.5-Sonnet & $\sim 175$ & \href{https://docs.anthropic.com/en/docs/about-claude/models}{https://docs.anthropic.com/en/docs/about-claude/models} \\
        \midrule
        Gemini & Gemini-1.5-Pro & $\sim 175$ & \href{https://ai.google.dev/gemini-api/docs/models/gemini?hl=en#gemini-1.5-pro}{https://ai.google.dev/gemini-api/docs/models/gemini?hl=en\#gemini-1.5-pro} \\
        \midrule
        \multirow{5}{*}{Deepseek} 
        & DeepSeek-v3 & 671 & \href{https://huggingface.co/deepseek-ai/DeepSeek-V3}{https://huggingface.co/deepseek-ai/DeepSeek-V3} \\
        & DeepSeek-R1-Distill-Qwen-7B & 7 & \href{https://huggingface.co/deepseek-ai/DeepSeek-R1-Distill-Qwen-7B}{https://huggingface.co/deepseek-ai/DeepSeek-R1-Distill-Qwen-7B} \\
        & DeepSeek-R1-Distill-Llama-8B & 8 & \href{https://huggingface.co/deepseek-ai/DeepSeek-R1-Distill-Llama-8B}{https://huggingface.co/deepseek-ai/DeepSeek-R1-Distill-Llama-8B} \\
        \midrule
        \multirow{3}{*}{Qwen} 
        & Qwen2.5-14B-Instruct & 14 & \href{https://huggingface.co/Qwen/Qwen2.5-14B-Instruct}{https://huggingface.co/Qwen/Qwen2.5-14B-Instruct} \\
        & Qwen2.5-7B-Instruct & 7 & \href{https://huggingface.co/Qwen/Qwen2.5-7B-Instruct}{https://huggingface.co/Qwen/Qwen2.5-7B-Instruct} \\
        \midrule
        GLM & GLM-4-9B-Chat & 9 & \href{https://huggingface.co/THUDM/glm-4-9b-chat}{https://huggingface.co/THUDM/glm-4-9b-chat} \\
        \midrule
        Yi & Yi-6B-Chat & 6 & \href{https://huggingface.co/01-ai/Yi-6B-Chat}{https://huggingface.co/01-ai/Yi-6B-Chat} \\
        \midrule
        LLAMA & Llama-3.1-8B-Instruct & 8 & \href{https://huggingface.co/meta-llama/Llama-3.1-8B-Instruct}{https://huggingface.co/meta-llama/Llama-3.1-8B-Instruct} \\
        \midrule
        Mistral & Mistral-7B-Instruct-v0.3 & 7 & \href{https://huggingface.co/mistralai/Mistral-7B-Instruct-v0.3}{https://huggingface.co/mistralai/Mistral-7B-Instruct-v0.3} \\
        \midrule
        Phi & Phi-3.5-mini-instruct & 3.8 & \href{https://huggingface.co/microsoft/Phi-3.5-mini-instruct}{https://huggingface.co/microsoft/Phi-3.5-mini-instruct} \\
        \bottomrule
    \end{tabular}
    \caption{Overview of Selected Large Language Models with Parameter Sizes and Reference Links}
    \label{tab:model-link}
\end{table*}

\section{Details of Prompts}
\label{sec:prompt}
\begin{figure*}[htbp]
	\centering
	\includegraphics[width=\textwidth]{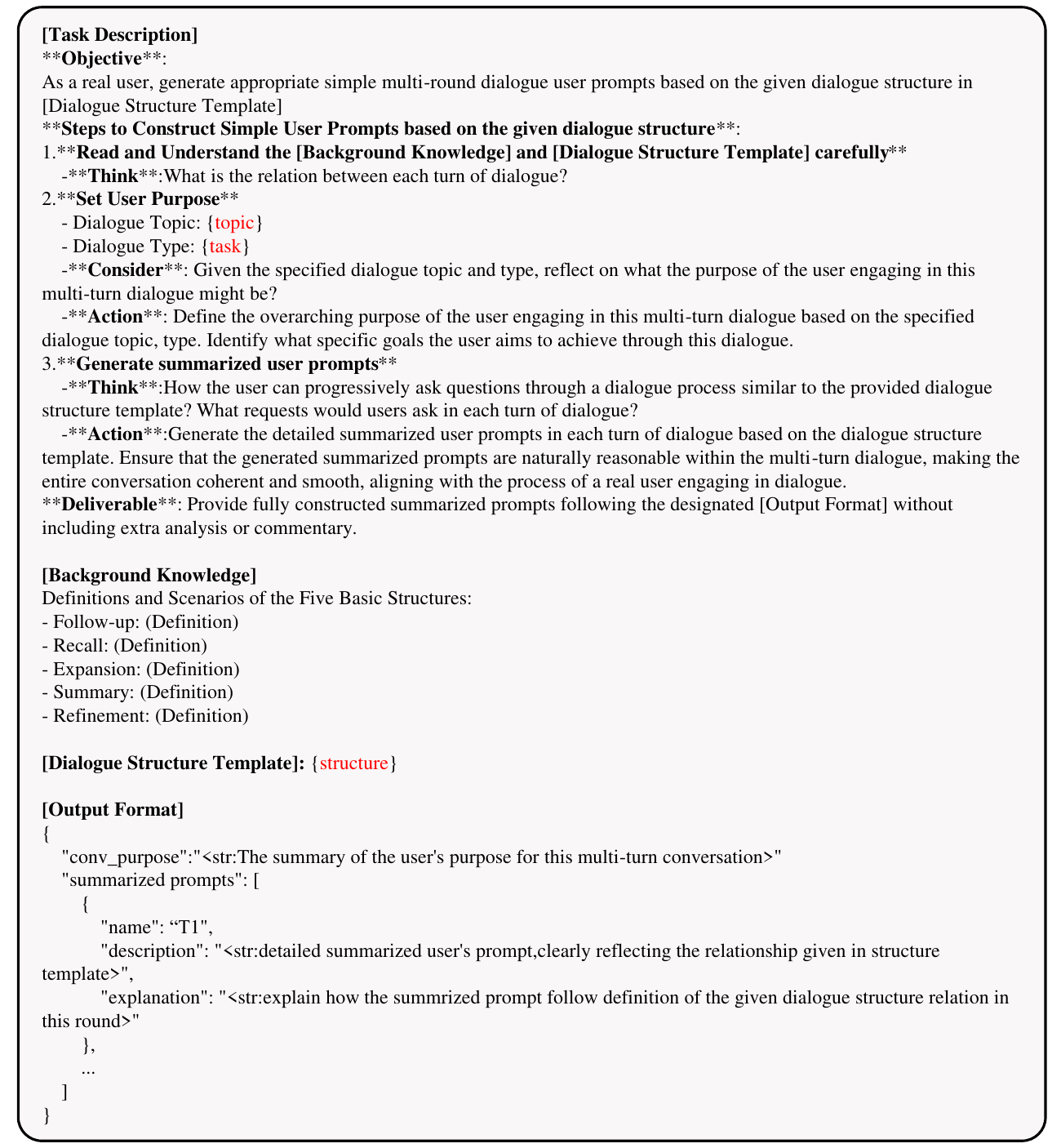}
	\caption{Intermediate Dialogue Plan Generation Template}
    \label{fig:imtermediate-prompt}
\end{figure*}

\begin{figure*}[htbp]
	\centering
	\includegraphics[width=\textwidth]{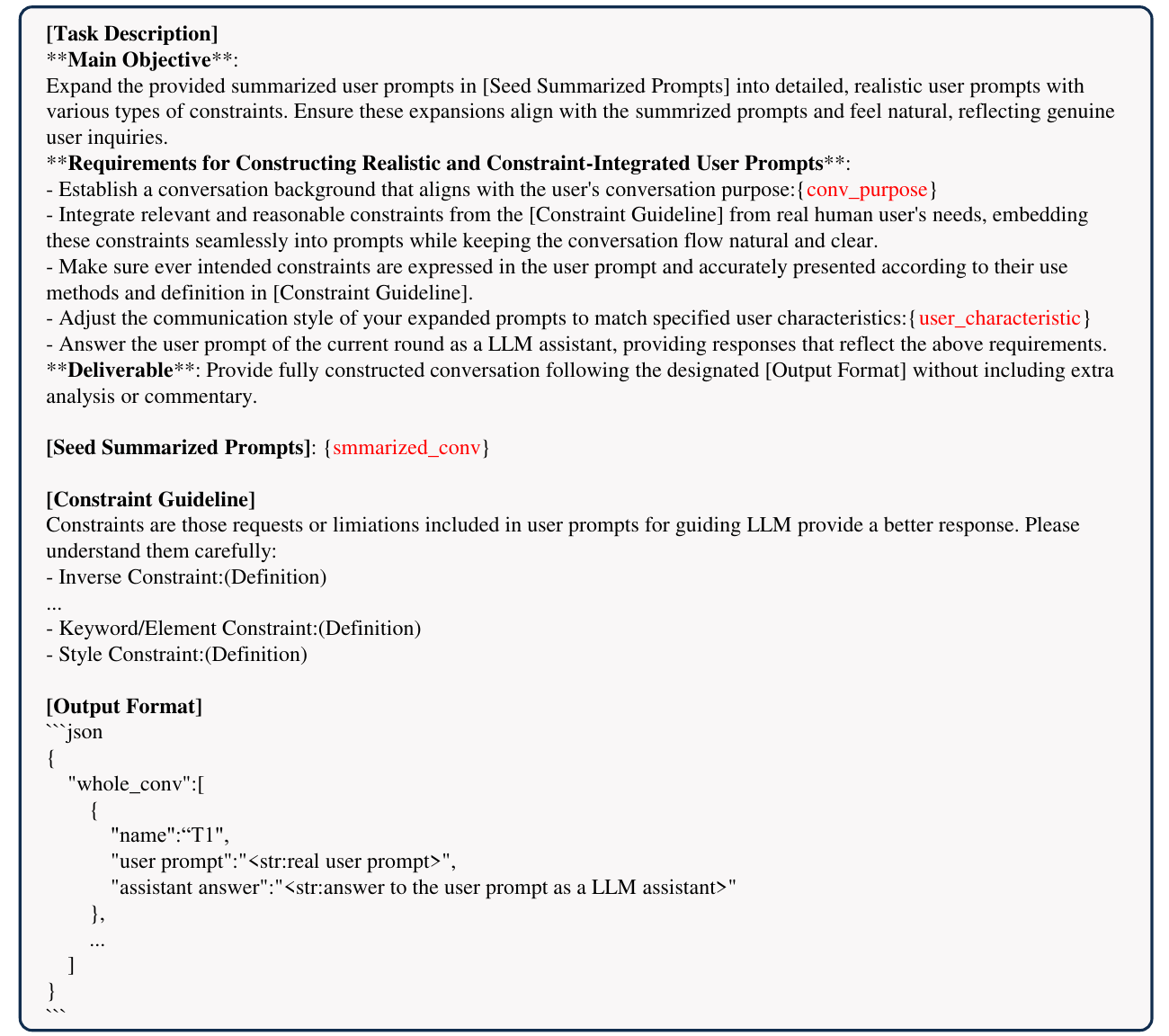}
	\caption{Complete Dialogue Generation Prompt Template}
    \label{fig:complete-prompt}
\end{figure*}

\begin{figure*}[htbp]
	\centering
	\includegraphics[width=\textwidth]{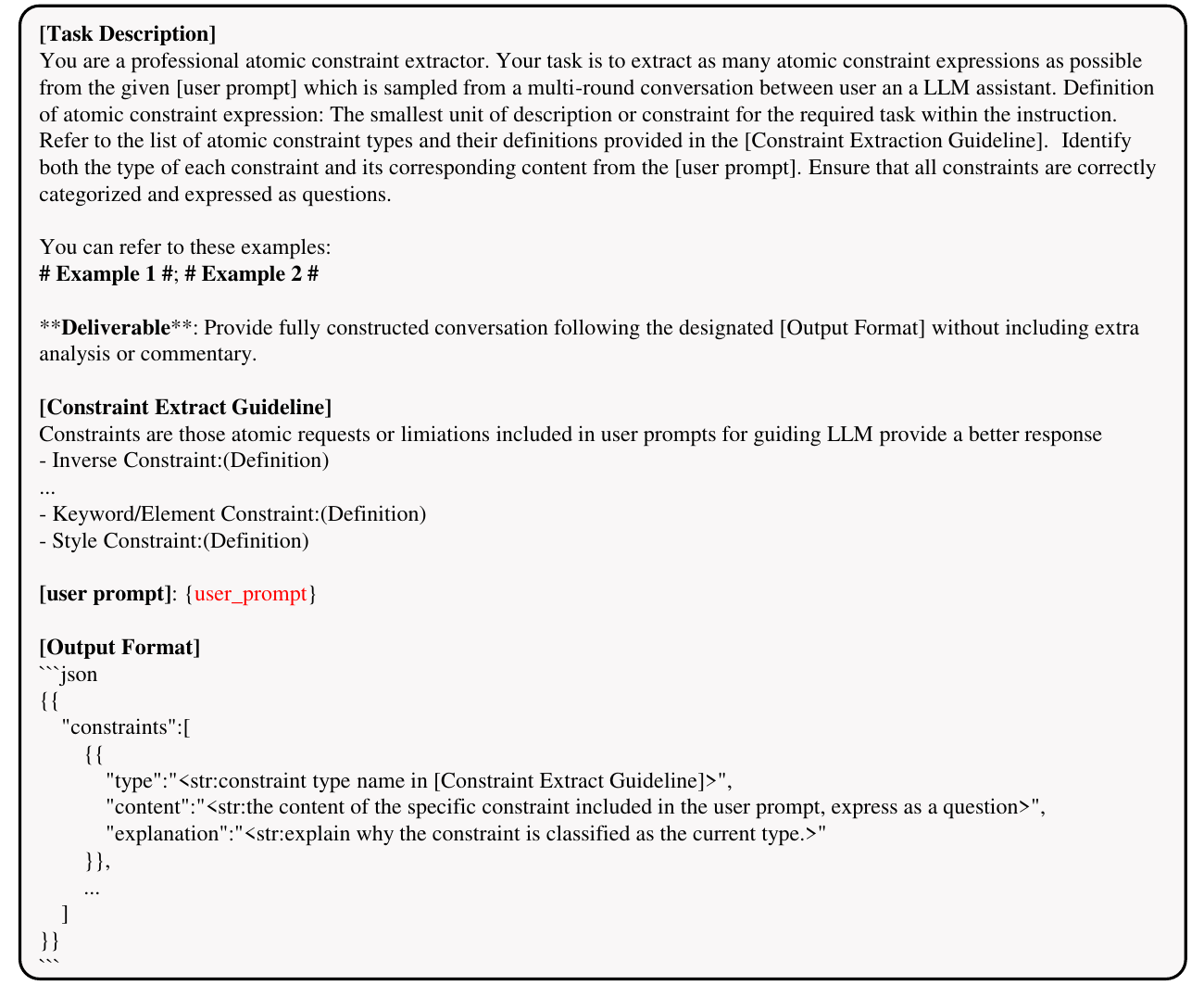}
	\caption{Constraint Extraction Prompt Template}
    \label{fig:extraction-prompt}
\end{figure*}

\begin{figure*}[htbp]
	\centering
	\includegraphics[width=\textwidth]{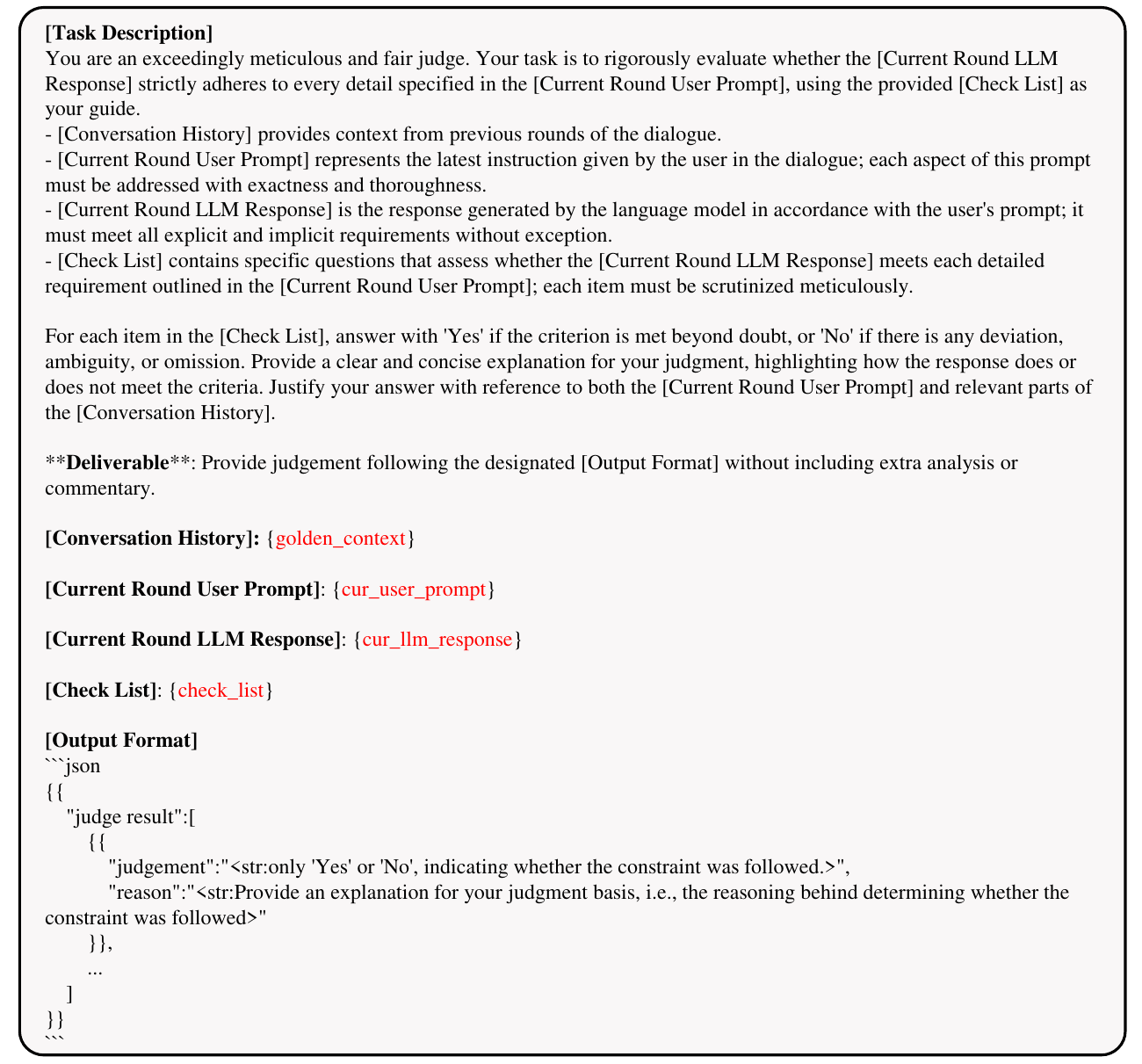}
	\caption{GPT-4o Evaluation Prompt Template}
    \label{fig:evaluation-prompt}
\end{figure*}

Figure~\ref{fig:imtermediate-prompt} to Figure~\ref{fig:evaluation-prompt} respectively illustrate the intermediate dialogue plan generation template, complete dialogue generation prompt template, constraint extraction prompt template, and GPT-4o evaluation prompt template used in our study.

\begin{figure*}[htbp]
	\centering
	\includegraphics[width=\textwidth]{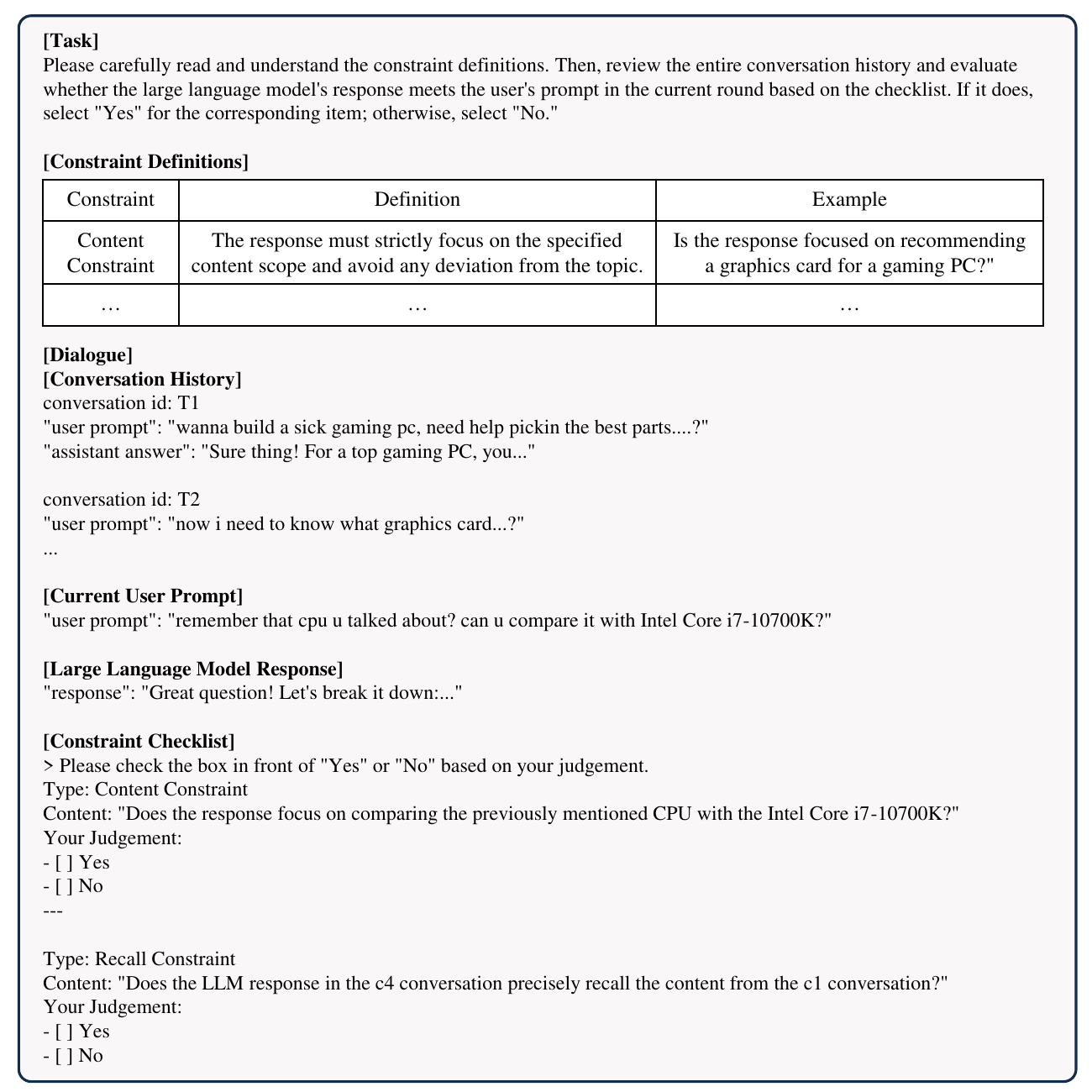}
	\caption{Task Requirements For Human Evaluators}
    \label{fig:task-requirement}
\end{figure*}

\end{document}